\documentclass[conference]{IEEEtran}
\IEEEoverridecommandlockouts
\usepackage{cite}
\usepackage{amsmath,amssymb,amsfonts}
\usepackage{algpseudocode}
\usepackage{algorithm}
\usepackage{amsmath}
\usepackage{amssymb}
\usepackage{graphicx}
\usepackage{textcomp}
\usepackage{booktabs}
\usepackage{url,todonotes,color}
\usepackage{array}
\usepackage{graphicx,subcaption}   
\usepackage{multirow}
 \usepackage{booktabs,balance}

\def\smb{\emph{Super Mario Bros}}

\def\BibTeX{{\rm B\kern-.05em{\sc i\kern-.025em b}\kern-.08em
    T\kern-.1667em\lower.7ex\hbox{E}\kern-.125emX}}

\begin{document}

\title{A Novel CNet-assisted Evolutionary Level Repairer and Its Applications to \smb
\thanks{T. Shu and Z. Wang contributed equally to this work.}
}

\author{
\IEEEauthorblockN{Tianye Shu\IEEEauthorrefmark{1}, Ziqi Wang\IEEEauthorrefmark{2}, Jialin Liu\IEEEauthorrefmark{3}, Xin Yao\IEEEauthorrefmark{4}}
\IEEEauthorblockA{Guangdong Provincial Key Laboratory of Brain-inspired Intelligent Computation\\ 
Department of Computer Science and Engineering\\ 
Southern University of Science and Technology\\ 
Shenzhen 518055, China\\
\IEEEauthorrefmark{1}11710101@mail.sustech.edu.cn,
\IEEEauthorrefmark{2}11710822@mail.sustech.edu.cn,
\IEEEauthorrefmark{3}liujl@sustech.edu.cn,
\IEEEauthorrefmark{4}xiny@sustech.edu.cn}}
\maketitle

\begin{abstract}
Applying latent variable evolution to game level design has become more and more popular as little human expert knowledge is required. However, defective levels with illegal patterns may be generated due to the violation of constraints for level design. A traditional way of repairing the defective levels is programming specific rule-based repairers to patch the flaw. However, programming these constraints is sometimes complex and not straightforward. An autonomous level repairer which is capable of learning the constraints is needed. In this paper, we propose a novel approach, CNet, to learn the probability distribution of tiles giving its surrounding tiles on a set of real levels, and then detect the illegal tiles in generated new levels. Then, an evolutionary repairer is designed to search for optimal replacement schemes equipped with a novel search space being constructed with the help of CNet and a novel heuristic function. The proposed approaches are proved to be effective in our case study of repairing GAN-generated and artificially destroyed levels of \emph{Super Mario Bros.} game. Our CNet-assisted evolutionary repairer can also be easily applied to other games of which the levels can be represented by a matrix of objects or tiles.
\end{abstract}


\section{Introduction}

    Procedural content generation (PCG) refers to the generation of game content (including rules, levels, maps, sound, background stories and so on) automatically without or with limited help of human designers~\cite{togelius2011procedural}. It has become a popular area in recent years. 
    With the development of video games, the scale of game production teams is also expanding. This leads to increasingly expensive development costs of game production. PCG techniques can reduce the workload of game designers or give designers inspiration. That means, for big game companies, PCG techniques can decrease the costs and shorten cycle for game production. On the other hand, PCG can also lower the technology and capital threshold of complex game development so that those small teams, even single-person developers, who have good ideas can realize their ideas with PCG techniques. What's more, some games need automatic content generation after their publication, including real-time content generation. Those games will be the biggest beneficiaries of development of PCG. 
    \par
    Mario AI framework is designed for the game AI competition for \emph{Super Mario} which includes a PCG track \cite{shaker20112010}. After 2010, some level generation methods for \emph{Super Mario} based on the Mario AI framework have been proposed (e.g.,~\cite{shaker2012evolving,dahlskog2012patterns,togelius2013patterns,dahlskog2014procedural,volz2018evolving,lucas2019tile}). Some of those methods are based on human designed grammars~\cite{shaker2012evolving} or patterns~\cite{dahlskog2012patterns, togelius2013patterns,dahlskog2014procedural}. Those methods have their own advantages respectively. However, they still need human designers to specify the behaviors of generators.
    \par
    In 2018, Volz \emph{et al.}~\cite{volz2018evolving} successfully applied latent variable evolution (LVE)~\cite{bontrager2018LVE} to level generation of \smb. They trained a Deep Convolution Generative Adversarial Networks (DCGAN) and use its generator to generate \smb~levels. Their approach is powerful, but they mentioned GAN will generate broken pipes sometimes in \cite{volz2018evolving}. This raised our interests in repairing the logic errors of levels generated by GAN.
    \par
    When the DCGAN generates the tiles to build a level, it cannot ``consider'' the surrounding neighbor tiles of each tile. We think this is the reason of broken pipes generated by GAN. Broken pipe is a microcosm of a stubborn problem of GAN-based method: there are usually some rules for game level design but the model cannot learn them due to its own network structure. One may think of using a Recurrent Neural Network (RNN) as the model of GAN \cite{mogren2016c}. RNNs can partially handle the broken pipes problem since they take some connection with neurons in the same layer. But the constraints of level may be complex, and the correctness of a tile may be influenced by the neighbor tiles in multiple directions. Besides, in terms of level generation, the network model of DCGAN may have some potential superiority comparing to RNNs. In fact, we have tried to train a GAN on a dataset of more levels in \smb~with 35 types of tiles in total. New types of tiles bring more constraints and cause more logic errors made by the GAN. Therefore, an automatic algorithm to repair levels with logic errors is desired.
    \par
    A straightforward way to automatically repair logic errors is writing a script to edit the level files based on some predefined rules. However, as mentioned previously, a significant advantage of LVE approach is that it can learn how to design levels without expert knowledge. If we need to design a repairing rule for each different game and each different type of errors, why not designing a pattern-based or grammar-based generator directly? We'd like to devise an approach to repair the levels generated by GAN without the help of human designers. In this work we focus on using AI techniques to (i) exam the generated level and detect the logic errors, and then (ii) repair the logic errors in levels without human designers' help.
    
    In this work, we come up with an approach to learn the level constraints from real levels and then repair the defective levels generated by GAN. This approach is inspired by the thinking process of human when repairing broken pipes for \smb~levels. When one sees a broken pipe, this person will detect several possibly error tiles by observing their surrounding tiles. And then, this person may think of some \emph{schemes} to replace those error tiles and analyze if this scheme is proper. After considering several replacement schemes, the best one and the resulted optimally repaired level will be determined. Our proposed approach imitates this process. First, we train a Multi-Layer Perceptron (MLP) model to judge whether a tile disobeys the constraint or not by taking its surrounding tiles as input. When repairing a defective level, we use the model to label the error tiles first, and then use a genetic algorithm (GA) to search for replacement schemes for those error tiles. In our case study, we have applied our approach to determining and repairing the broken pipes in \smb~levels generated by MarioGAN~\cite{volz2018evolving} and our experimental study has validated its effectiveness.
   
    The remainder of this paper is organized as follows. Some related work on PCG and Mario level generation are presented in Section \ref{sec:background}. 
    Section \ref{sec:rulenet} presents our proposed CNet for detecting error tiles in generated levels and recommending alternatives for repairing.
    Our novel evolutionary repairer is presented in Section \ref{sec:enhancedga}. The experimental studies on these two approaches are described in Sections \ref{sec:rulenetxps} and \ref{sec:gaxps}, respectively.  
    Section \ref{sec:conclusion} concludes and discusses future work.
    

\section{Background}\label{sec:background}
    Section \ref{sec:mariopcg} reviews the level generation methods for \smb. Section \ref{sec:mariogan} presents particularly the MarioGAN framework. The other ingredient, genetic algorithm, is briefly introduced in Section \ref{sec:ga}.

    \subsection{Mario Level Generation}\label{sec:mariopcg}
    Level generation is an important part in PCG. The 2010 Mario AI Championship held an academic PCG competition \cite{shaker20112010} which the participants need submit a generator for \smb~level. This competition touched off a series of studies on Mario level generation. Shaker \emph{et al.} used Multi-Layer Perceptron (MLP) models to predict the players' preference \cite{shaker2010towards}, and applied grammar evolution to generate personalized level by optimizing the predicted value of their model \cite{shaker2012evolving}. Steve \emph{et al.} designed many patterns for Mario levels\cite{dahlskog2012patterns} and using patterns as objectives and generate levels by evolutionary algorithms (EA) \cite{dahlskog2014procedural}. Those methods belong to search-based methods\cite{yannakakis2018artificial}, they usually need some expert knowledge. 
    \par
    In recent years, Procedural Content Generation via Machine Learning (PCGML) \cite{summerville2018procedural} has been more and more popular since it needs little expert knowledge of target games. Those methods usually represent the levels as a matrix of tiles, just like image, then generate the tiles one by one using some machine learning technologies for image processing. Snodgrass and Ontañón generated Mario levels with Markov chains~\cite{snodgrass2014experiments}. In 2016, Jain \emph{et al.} applied autoencoders on level generation, repair, and recognition~\cite{jain2016autoencoders}. They focused on repairing unplayable levels but not logic errors of levels~\cite{jain2016autoencoders}. Summerville \emph{et al.} trained LSTM to learn level distribution and generate new levels~ \cite{summerville2016super}. Gutierrez and Schrum~\cite{gutierrez2020zeldagan} combined a GAN method and a graph grammar method to generate levels for The Legend of Zelda.
    Schrum \emph{et al.}~\cite{schrum2020cppn2gan} used Compositional Pattern Producing Networks (CPPNs) to combine Zelda level segments generated by GANs into complete levels. \cite{schrum2020interactive} presented an interactive level design tool which allows the human designer to edit latent vectors of the trained generative models to generate diverse levels that can be represented as two-dimensional arrays of tiles.
    In 2019, Lucas and Volz used Kullback-Leibler (KL) to analyze game levels, proposed a level generation method called Evolution with Tile Pattern KL Divergence (ETPKLDiv) \cite{lucas2019tile}. Broken pipes also occur in the generated levels illustrated in Figure 6 of~\cite{lucas2019tile}.
    
    There are many methods for Mario level generation, each method has its own advantages. However, broken pipes problem is a common problem for some Mario level generators, especially those who need little expert knowledge.
    
    
    

   
    \subsection{MarioGAN}\label{sec:mariogan}
    DCGAN was successfully applied to level generation of \smb~in 2018 \cite{volz2018evolving}. In \cite{volz2018evolving}, a level was represented by a two-dimensional array of tile codes, while each code referred to one of the 10 different tile types. The Video Game Level Corpus (VGLC)~\cite{summerville2016vglc} encoding is to used represent the levels. Table \ref{tab:tiles} shows the tiles, their corresponding VGLC encoding and the mapped values. We use the same encoding in our work. In their work, only one level of \smb~was used to train a Generative Adversarial Network (GAN), and then a Covariance Matrix Adaptation Evolution Strategy (CMA-ES) was used to search in the latent space to further optimize the generated levels~\cite{volz2018evolving}. In the optimization phase, an AI agent was used to test the generated levels. This framework also called Exploratory Latent Search GAN (ELSGAN) by Lucas and Volz \cite{lucas2019tile}. One of the most attractive qualities of ELSGAN is that it can learn how to design levels automatically without predefined generation rules, grammars or patterns for the specific game. That's to say, it may be applied in any games whose level can be represented as a two-dimensional array of discrete values without the knowledge of the specific game. However, this framework cannot learn the combination constraints of levels which leads to some generated levels with logic errors. Taken the \smb~as an example, broken pipes may appear in the levels generated by GAN. 
    
\begin{table}[htbp]
\centering
\caption{\label{tab:tiles}Tile types and encoding for generating Mario levels used in~\cite{volz2018evolving,lucas2019tile,schrum2020interactive,schrum2020cppn2gan} and this paper.}
    \begin{tabular}{cccc}
    \hline
    Tile type & Symbol & Identity & Visualization\\
    \hline 
    Solid/Ground & X & 0 & \includegraphics[scale=0.5]{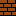}\\
    Breakable & S & 1 & \includegraphics[scale=0.5]{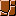}\\
    Empty (passable) & - & 2 & 
    \includegraphics[scale=0.5]{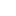}\\
    Full question block & ? & 3 & \includegraphics[scale=0.5]{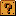}\\
    Empty question block & Q & 4 & \includegraphics[scale=0.5]{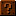}\\
    Enemy & E & 5 & \includegraphics[scale=0.5]{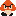}\\
    Top-left pipe & \textless & 6 & \includegraphics[scale=0.5]{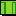}\\
    Top-right pipe & \textgreater & 7 & \includegraphics[scale=0.5]{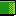}\\
    Left pipe & [ & 8 & \includegraphics[scale=0.5]{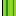}\\
    Right pipe & ] & 9 & \includegraphics[scale=0.5]{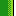}\\
    Coin & o & 10 & \includegraphics[scale=0.5]{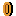}\\
    \hline
    \end{tabular}
\end{table}
    
    \subsection{Genetic Algorithm}\label{sec:ga}
    Evolutionary computation techniques have been widely applied in optimizing game-playing strategies and designing games, in particular tuning game AI agents and evolving game skill-depth, such as in \cite{sironi2018self,liu2017evolving}.
    Genetic Algorithm (GA)~\cite{holland1992genetic} is a family of evolutionary algorithms which simulates the process of the inheritance and evolution of the creature. GA is suitable for approximately solving black-box optimization problems in discrete domain. At each generation, GA applies some search operators to generate new individuals (solutions) and select the best ones to survive for the next generation. Two of the major challenges of GA are the design of representation and fitness function. In this work, we have designed a novel fitness function for repairing game levels and defined a narrowed search space but with diversities to accelerate the search.
   \section{CNet: a novel automatic quality controller}\label{sec:rulenet}
   
    \begin{figure}[htbp]
        \centering
        \includegraphics[width=1\columnwidth]{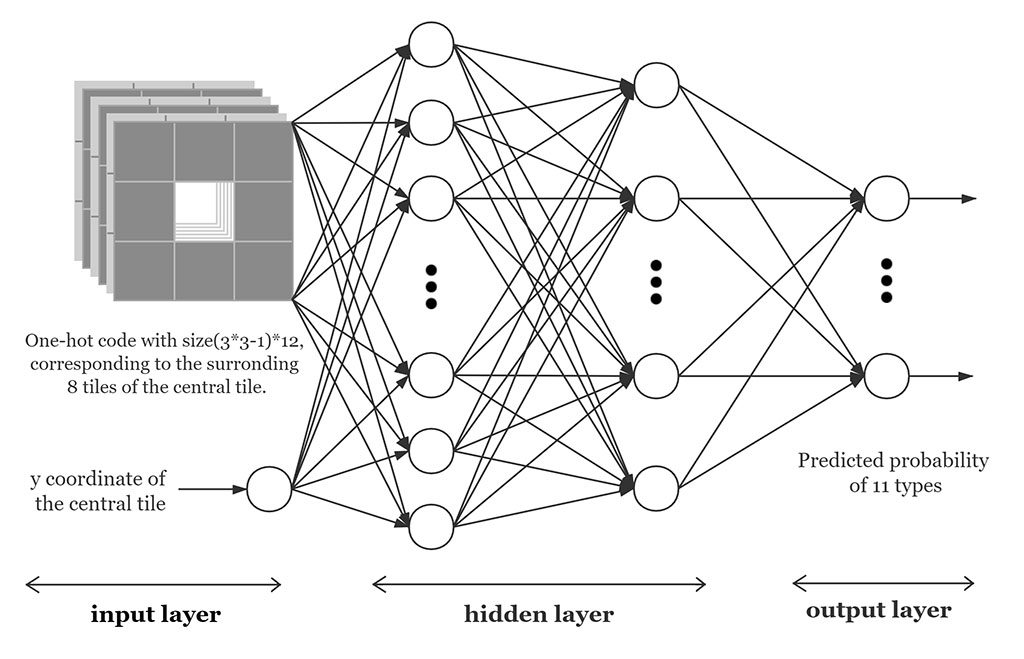}
        \caption{Illustration of the CNet.}
        \label{fig:ruleNet}
    \end{figure}
    
    We designed a MLP model, named as \emph{CNet}, to learn the constraints of pipes in \smb~levels and check the levels generated by GAN. 
            
    Before introducing the CNet in details, we present the definitions used in this paper. For any position $(i,j)$ in a level, we consider a tile grid of size $3\times 3$ with $(i,j)$ as the center. A \textbf{combination} is defined as a sequence of the height of the central tile and the 9 tile types in the $3\times 3$ grid.
    A \textbf{surrounding} \textbf{info} is defined as a sequence of the height of the central tile and the tile types for its 8 surrounding tiles.
    For each position and tile type, the surrounding info which appears in the training data is called \textbf{true surrounding} info, while the ones do not appear in the training data are called \textbf{fake surrounding info}. 
    For instance, given a combination $(13,2,2,5,0,6,7,0,8,9)$, then the type of the central tile is $6$ (top-left pipe) and its height is $13$. The corresponding surrounding info is $(13,2,2,5,0,7,0,8,9)$.
    For a given true surrounding info, a tile is called \textbf{legal tile} if there exist a combination in training data which contains the surrounding info and this tile as center tile. Otherwise, this tile is called \textbf{illegal tile}. For example, giving $(13,2,2,5,0,7,0,8,9)$, the ``6'' is a legal tile. The ``0'' is an illegal tile since the combination  $(13,2,2,5,0,0,7,0,8,9)$ does not appear in the training data.
    
   \subsection{CNet}\label{sec:rulenetdesign}
    The CNet executes as follows. For each of the tiles of a given generated level being checked, the CNet takes its $y$-coordinate and its surrounding tiles (i.e., a surrounding info) as input, as we assume that the vertical position of the tile may be significant to judge its rationality in this game, and outputs the predicted possible types of this tile being inspected. The process is illustrated in Fig. \ref{fig:ruleNet}.  
    The input layer can be seen as a hollow cube of $(3*3-1)*12$ size plus one single input. Width and height of the cube correspond to the relative position of the center and the 12 channels correspond to the 12 tile types including the outer type which refers to
    the periphery of the level. There are 2 full connected hidden layers in CNet. The first layer has 100 neurons and the second one has 50 neurons. Softmax is used at the output layer and CrossEntropy loss is used. The output layer gives the probability distribution over the tile types listed in Table \ref{tab:tiles}, of which the sum is $1$.

    \subsection{CNet as a tile inspector}\label{sec:inspector}
    For each position $(i,j)$ in the generated level being checked, let $t$ denote the present tile type. the CNet takes into its surrounding tiles as input and output the probabilities of having different tile types at $(i,j)$. 
    If the probability of having the tile type $t$ at position $(i,j)$, denoted as $P(T_{i,j}=t)$, is under a given threshold $\theta$, it is a \textbf{wrong tile}.
    \footnote{An ``error tile'' is a tile that is determined as wrong by human while a ``wrong tile'' is a special definition for CNet (Section \ref{sec:rulenet}).\label{error_tile}}
    \subsection{CNet as a tile recommender}\label{sec:recommender}
    When the CNet determines a wrong tile, it can also recommend tiles for this position.
    To do so, a \emph{truncation} step is introduced to detect the less possible tile types for classification. For any position, the tile type with a probability above the threshold $\theta$ is defined as a \textbf{candidate type} for this position. There could be one or more candidate types at the tile being checked. Therefore, a wrong tile can also be defined as the tile that is not filled with a candidate type.
    
       \subsection{Experiments and results}\label{sec:rulenetxps}
        We define \textbf{unstable tiles} as the tiles that have more than one candidate types of a level $l$ being repaired. The set of unstable tiles in a level $l$ is denoted as $\mathcal{U}(l)$. $\mathcal{C}_{i,j}^l$ denotes the set of candidate type at position $(i,j)$ of level $l$. An \textbf{unstable value}, $UV(l)$, is defined as the sum of number of candidate types for all unstable tiles in it, formalized as follows:
  \begin{equation}\label{eq:uv}
      UV(l) = \sum \limits_{(i,j) \in \mathcal{U}(l)} \vert \mathcal{C}_{i,j}^l \vert.
  \end{equation} 
    To test our CNet and evolutionary repairer, we have designed three experiments and corresponding test data sets to answer the following three questions:
    \begin{itemize}
        \item  How does CNet perform in recommending legal tiles and eliminating illegal tiles?
        \item 
        How would CNet perform giving a fake surrounding info?
        \item When the surrounding info is fake, thus it does not appear in training data, will the amount of unstable tiles and unstable values increase?
    \end{itemize}
    
    In all the experiments, seventeen levels from the Mario AI framework are used as training data. The CNet has been training for $4,000$ epoch using all non-repeated $3*3$ grids from the training data. The truncation threshold, $\theta$, is set as 0.05 in all experiments.
      
        \subsubsection{CNet classifies excellently legal tiles and illegal tiles}

    To test the performance of CNet in recommending legal tiles and eliminating illegal tiles, we have designed two tests and generated two test data sets. 
    
    The first test data set contains all combinations which appear in the training data. It is aimed to test whether the set of candidate type provided by our CNet contains the legal tile's type giving true surrounding info.
    The second test data set contains all combinations which do not appear in the training data, but there is no fake surrounding info. It is aimed to test whether illegal tiles will be detected by our CNet giving true surrounding info. These results are categorized by the type of central tile.
    \begin{table}[htbp]
    \centering
    \setlength{\tabcolsep}{.9mm}{
    \caption{Results of identifying legal tiles. ``\#Elm'' refers to the number of eliminated tiles.}
    \label{tab:neg}
        \begin{tabular}{c|r|r|r|r|r|r|r|r|r}
            \toprule
            \textbf{Tile} &
            \multirow{2}{*}{\textbf{Total}} &
            \multicolumn{2}{c|}{\textbf{True}}    &
            \multicolumn{2}{c|}{\textbf{Fake1}}    &
            \multicolumn{2}{c|}{\textbf{Fake2}}    &
            \multicolumn{2}{c}{\textbf{Fake3}} \cr
            \textbf{Type} & & \#Elm & Rate & \#Elm & Rate & \#Elm & Rate & \#Elm & Rate
            \cr\midrule
            0	& 76	& 0	& 0.0\%	& 8	& 10.5\%	& 10	& 13.2\%	& 13	& 17.1\%
            \cr\hline
            1	& 23	& 0	& 0.0\%	& 1	& 4.4\%	& 4	& 17.4\%	& 7	& 30.4\%
            \cr\hline
            2	& 294	& 0	& 0.0\%	& 14	& 4.8\%	& 20	& 6.8\%	& 28	& 9.5\%
            \cr\hline
            3	& 2	& 0	& 0.0\%	& 0	& 0.0\%	& 0	& 0.0\%	& 1	& 50.0\%
            \cr\hline
            4	& 13	& 0	& 0.0\%	& 0	& 0.0\%	& 4	& 30.8\%	& 5	& 38.5\%
            \cr\hline
            5	& 20	& 0	& 0.0\%	& 3	& 15.0\%	& 4	& 20.0\%	& 8	& 40.0\%
            \cr\hline
            6	& 33	& 0	& 0.0\%	& 5	& 15.2\%	& 5	& 15.2\%	& 14	& 42.4\%
            \cr\hline
            7	& 34	& 0	& 0.0\%	& 1	& 2.9\%	& 5	& 14.7\%	& 5	& 14.7\%
            \cr\hline
            8	& 69	& 0	& 0.0\%	& 5	& 7.3\%	& 7	& 10.1\%	& 8	& 11.6\%
            \cr\hline
            9	& 72	& 0	& 0.0\%	& 0	& 0.0\%	& 3	& 4.2\%	& 7	& 9.7\%
            \cr\bottomrule
        \end{tabular}   }
    \end{table}

    The ``True'' column of Table \ref{tab:neg} summaries the results of the first test. 
    Note that all the tiles are legal in this test, thus there is no negative data. For each central tile, if the candidate types recommended by our CNet contains the legal tile, then it is considered as a correct classification. The reason is that giving a true surrounding info, there could be more than one legal central tile. Results show that the candidate tiles always contains the legal tiles of which the codes are 6, 7, 8 and 9. They are the codes for composing pipes. It means that our CNet is able to learn the combination constraints of different types of pipes. But for tiles around pipes, such as the ground tiles (``0''), breakable tiles (``1'') and monster tiles (``5''), legal tiles are sometimes eliminated from candidate types. Especially, the full question tile (``3'') is always wrongly eliminated from the set of candidate types. Actually, the total number of combinations which have a full question tile as central tile is $2$. It is possible that the CNet did not learn this constraint well due to the very limited amount of data. 
    The ``True'' column of Table \ref{tab:pos} gives results of the second test. When giving true surrounding info, our CNet successfully detected correctly most of the illegal tiles.\par
    In conclusion, CNet performs excellently in recommending legal tiles and eliminating illegal tiles.
    \begin{table}[thbp]
    \centering
    \setlength{\tabcolsep}{.9mm}{
    \caption{Detecting illegal tiles test. ``\#Det'' refers to the number of detected tiles.}
    \label{tab:pos}
        \begin{tabular}{c|r|r|r|r|r|r|r|r|r}
            \toprule
            \textbf{Tile} &
            \multirow{2}{*}{\textbf{Total}} &
            \multicolumn{2}{c|}{\textbf{True}}    &
            \multicolumn{2}{c|}{\textbf{Fake1}}    &
            \multicolumn{2}{c|}{\textbf{Fake2}}    &
            \multicolumn{2}{c}{\textbf{Fake3}} 
            \cr
            \textbf{Type} & & \#Det & Rate & \#Det & Rate & \#Det & Rate & \#Det & Rate
            \cr\midrule
            0	& 553	& 552	& 99.8\%	& 520	& 94.0\%	& 514	& 92.9\%	& 491	& 88.8\%
            \cr\hline
            1	& 607	& 602	& 99.2\%	& 591	& 97.4\%	& 582	& 95.9\%	& 576	& 94.9\%
            \cr\hline
            2	& 331	& 327	& 98.8\%	& 296	& 89.4\%	& 281	& 84.9\%	& 266	& 80.4\%
            \cr\hline
            3	& 627	& 627	& 100.0\%	& 623	& 99.4\%	& 621	& 99.0\%	& 612	& 97.6\%
            \cr\hline
            4	& 616	& 616	& 100.0\%	& 605	& 98.2\%	& 595	& 96.6\%	& 590	& 95.8\%
            \cr\hline
            5	& 609	& 607	& 99.7\%	& 600	& 98.5\%	& 596	& 97.9\%	& 584	& 95.9\%
            \cr\hline
            6	& 597	& 597	& 100.0\%	& 593	& 99.3\%	& 594	& 99.5\%	& 590	& 98.8\%
            \cr\hline
            7	& 596	& 596	& 100.0\%	& 590	& 99.0\%	& 583	& 97.8\%	& 572	& 96.0\%
            \cr\hline
            8	& 560	& 558	& 99.6\%	& 555	& 99.1\%	& 549	& 98.0\%	& 547	& 97.7\%
            \cr\hline
            9	& 556	& 556	& 100.0\%	& 546	& 98.2\%	& 530	& 95.3\%	& 526	& 94.6\%
            \cr\hline
            10	& 629	& 629	& 100.0\%	& 629	& 100.0\%	& 629	& 100.0\%	& 629	& 100.0\%
            \cr\bottomrule
        \end{tabular}   }
    \end{table}

    \subsubsection{Fake surrounding info made classification harder}
    True surrounding info may become fake when there is one or more error tiles in it. In the generated levels, multiple error tiles may be adjacent. When repairing one of them, the surrounding info is fake. To test the performance of CNet giving fake surrounding info, we generate test data on basis of the test data in the previous experiments. We randomly change the type of one, two or three tiles in the true surrounding info to make it fake.

    Firstly, we want to test whether legal tiles of true surrounding info still contains in the candidate types of fake surrounding info. The columns ``Fake 1'', ``Fake 2'', and ``Fake 3''  of Table \ref{tab:neg} show the results of using the data sets with one, two and three randomly changed tiles in combinations. When the surrounding info is fake, the number of cases of eliminating legal tile wrongly increases substantially.
    
    Secondly, we want to test whether illegal tiles of true surrounding info will be detected from the candidate tiles of fake surrounding info. Table \ref{tab:pos} shows the result. When the surrounding info is fake, the elimination rate decreases. This means that more illegal tiles are contained in candidate tiles.
    
    In conclusion, fake surrounding info makes the classification harder.

     \subsubsection{The Unstable Value is informative}
    In this experiment, we want to answer the third research question and check if the unstable value and unstable tiles can assist with the identification of broken areas in a level. To do so, we have generated three test data sets by editing the original training data.
    Giving a surrounding info as input, the trained CNet outputs the candidate types for the center cell. Then the unstable value (Eq. \eqref{eq:uv}) of this center tile is calculated to determine whether it is an unstable tile or not. Then we randomly change the type of one, two or three tiles in the surrounding info to generate a new tile probably with the fake surrounding info. Again, the unstable values and amount of unstable tiles are calculated for each case.
    We also calculate the original unstable values without classification of CNet for the training data as reference. In the training data, if any two different tile types have the same surrounding info, we say that this surrounding info corresponds to an unstable tile and the unstable value is at least two.
    \par
    Table \ref{tab:stable} shows the results.
    We can see that the amount of unstable tiles in the ``True'' column is greater than the amount in the ``Original'' column. It reflects that CNet mixed illegal tiles into the candidate types for some surrounding info. The ``Fake1'', ``Fake2'', ``Fake3'' columns have greater unstable values and larger amount of unstable tiles than the ``True'' column. This implies that fake surrounding info is linked with unstable tile. CNet may not judge well giving fake surrounding info. Therefore, the possibility is shared, making more tile types into the set of Candidate Types. So the unstable value and the amount of unstable tiles increase. When repairing levels, we can use the unstable value as an indicator. Smaller unstable value means less fake surrounding info in levels. Less fake surrounding info means fewer broken areas in levels.
    
     In conclusion, unstable value indicates the amount of fake surrounding Info. And unstable value can assist with the identification of broken areas in a level.
    
    \begin{table}[tp]
    \centering
    \setlength{\tabcolsep}{1.5mm}{
    \caption{Stable test. Each test set has 627 tiles in total. ``$|U|$'' refers to the number of unstable tiles and ``$UV$'' is the unstable value.}
    \label{tab:stable}
        \begin{tabular}{c|c|c|c|c|c|c|c|c|c|c}
            \toprule
            \multirow{2}{*}{\textbf{Set}}  &
            \multicolumn{2}{c|}{\textbf{Original}}    &
            \multicolumn{2}{c|}{\textbf{True}}    &
            \multicolumn{2}{c|}{\textbf{Fake1}}    &
            \multicolumn{2}{c|}{\textbf{Fake2}}    &
            \multicolumn{2}{c}{\textbf{Fake3}} 
            \cr
            & $|U|$ & $UV$ & $|U|$ & $UV$ & $|U|$ & $UV$ & $|U|$ & $UV$ & $|U|$ & $UV$
            \cr\midrule
            1	
            & 9     & 20    & 12	& 24	& 50	& 109	& 62	& 136	& 94	& 204
            \cr\hline
            2	
            & -- -- & -- -- & -- --	& -- -- & 49	& 103	& 87	& 197	& 100	& 214
            \cr\hline
            3	
            & -- -- & -- -- & -- --	& -- --	& 60   & 131   & 87	& 186	& 99	& 220
            \cr\bottomrule
        \end{tabular} 
        }
    \end{table}

    \subsection{Discussion}
    The CNet performs well in detecting the illegal tiles and recommending alternative for fixing the wrong tiles.
    To fix a wrong tile, it seems to be straightforward to directly replace it by the candidate tile type with the highest probability classified by the CNet. However, our primary experimental study (omitted due to the length limit) shows that this simple repairing operation, winner-takes-all, performed poorly on the Mario levels generated by GAN. There are mainly two reasons. First, an error tile will influence the decision of its surrounding tiles. Secondly, our CNet does not always make perfect classification.
    To offset the imperfection of CNet, we propose to use a genetic algorithm to search in the combination space and design some novel heuristic functions to direct the search.
 
    \section{Evolutionary repairers with a novel heuristic}\label{sec:enhancedga}
    As a level is a two-dimensional array of discrete values, we use a genetic algorithm (GA) to search for good replacement schemes (thus, combinations) in the original level to repair the broken areas. Although we're actually optimizing a replacement scheme, our work treats the solution of the search as a complete level for convenience. The encoding of the solution is a dictionary whose key is the position to be replaced and value is the alternative type.    
   
    \subsection{A narrowed search space with diversities}
    \label{sec:solutionspace}
    Defining the search space is not trivial. Primary experimental study using only the wrong tiles as the solution space showed that, sometimes, only replacing those tiles could not lead to a correct level at all. However, we still want to narrow the solution space as much as possible to guarantee the efficiency of our algorithm and keep the features from the original levels as much as possible. Therefore, we define the solution space as the set of all the error tiles and all the tiles that have alternatives. We use $\mathcal{S}_l$ to denote the set of wrong tiles and unstable tiles in a training level $l$. 

\subsection{Search operators}
    Algorithm \ref{algo:operators} presents our search operators. The crossover operator aims to combine good replacement schemes randomly to get better solutions. For a given solution $x$, our mutation operator generates a new solution $x'$ by replacing the tiles in $\mathcal{S}_l$ with types in candidate types of original level $l$ at position $(i,j)$. We design this operator to add new individuals into the population. The repair operator changes the types of the unstable tiles to a randomly selected candidate type. We design it to search in the local space. 
    Our main process (Algorithm \ref{algo:evolution}) performs crossover, selection, mutation, selection, repair and selection in turn at each generation. For the selection operator, we just select the best $n$ individuals.
    
    \begin{algorithm}[htbp]
\caption{\label{algo:operators}Search operators for repairing a level $l$. $\mathcal{S}_l$ denotes the set of wrong tiles and unstable tiles in $l$, represented by their positions. $x(i,j)$ denotes the element at position $(i,j)$ of 2D array $x$.}
\begin{algorithmic}[1]
\Function{Crossover}{$x_1, x_2$}
\State $x'_1 \gets x_1$, $x'_2 \gets x_2$
        \For{$(i,j) \in \mathcal{S}_l$}
            \If{$UniformRandom(0,1) \leq 0.5$}
                \State Swap $x'_1(i,j)$ and $x'_2(i,j)$
            \EndIf
        \EndFor
        \State\Return $x'_1$, $x'_2$
\EndFunction
\Statex
\Function{Mutation}{$p_{m1}$, $x$}  
        \For{$(i,j) \in \mathcal{S}_l$}
            \If{$UniformRandom(0,1) \leq p_{m1}$ and $x(i,j)$ is an unstable tile}
                \State $x(i,j) \gets$ randomly select in $\mathcal{C}^{x}_{i,j}$
            \EndIf
        \EndFor
\EndFunction
\Statex
\Function{Repair}{$p_r$, $x$}
        \For{$(i,j) \in \mathcal{S}_l$}
            \If{$UniformRandom(0,1) \leq p_{r}$ and $x(i,j)$ is a wrong tile}
                \State $x(i,j) \gets$ randomly select in $\mathcal{C}^x_{i,j}$
            \EndIf
        \EndFor
\EndFunction
\end{algorithmic}
\end{algorithm}

    \begin{algorithm}[htbp]
	\caption{\label{algo:evolution}Evolution process for repairing a level $l$.}
	\begin{algorithmic}[1]
        \State \textbf{Input:} $p_{m0} \gets 0.8$, $p_{m1} \gets \frac{1}{|\mathcal{S}_l|}$, $p_r\gets 0.3$, $n\gets 20$, $RRT_m\gets 4$
        \State Initialize population $\mathcal{X}$ (cf. Section \ref{sec:gaxps})
        \While{time not elapsed}
            \State Evaluate fitness of all $x$ in $\mathcal{X}$
            \State Calculate $P_x$ for all $x$ by $\frac{descendingRank(x)}{\sum_{x \in \mathcal{X}} descendingRank(x_i)}$
            \State$\mathcal{X'}\gets\emptyset$
            \While{$|\mathcal{X'}|<n$}
                \State Choose different $x_i,x_j$ from $\mathcal{X}$ according to probability $P_x$ ($x \in \mathcal{X}$) 
                \State $x_1,x_2 \gets$\Call{Crossover}{$x_i, x_j$}
                \State Add $x_1,x_2$ into $\mathcal{X}'$
            \EndWhile
            \For{$x$ in $\mathcal{X'}$}
            \If{$UniformRandom(0,1) \leq p_{m0}$}
  \State\Call{Mutation}{$p_{m1}$, $x$}
            \EndIf
            \EndFor
            \For{$x$ in $\mathcal{X'}$}
                \State\Call{Repair}{$p_r$, $x$}
            \EndFor
            \State Evaluate fitness of all $x$ in $\mathcal{X'}$
            \State $\mathcal{X} \gets $ The best $n$ individuals of $\mathcal{X} \cup \mathcal{X}'$ selected using Round-robin Tournament.
        \State Update the best individual
        \EndWhile
        \State\Return{the best individual}
        
	\end{algorithmic}
\end{algorithm}

\subsection{Fitness design}\label{sec:heuristic}
    We want to minimize the number of errors in the final solution. Besides, we want to keep the features from the original levels as much as possible. So we minimize the number of Replaced Tiles, $\vert \mathcal{R}(x) \vert$. Additionally, the Unstable Value is used as a third item to assist with repairing. When the amount of fake surrounding info decreases, the performance of CNet will be better. 
    As a result, a weighted sum of the above three metric is used as the fitness function in our work, formalized as follows:
    \begin{equation}\label{eq:fitness}
    F(x) = \omega_1 \cdot \vert \mathcal{W}(x) \vert + \omega_2  \cdot \vert \mathcal{R}(x) \vert +  \omega_3 \cdot UV(x),
    \end{equation}
    where $ \mathcal{W}(x)$ is the set of wrong tiles and $\omega_i$ are coefficients to be set, $i\in \{1,2,3\}$.
   
\subsection{Experimental setting and results}\label{sec:gaxps}
    \begin{figure}[htbp]
        \centering
        \begin{subfigure}{.5\textwidth}
            \centering
            \includegraphics[width=1\textwidth]{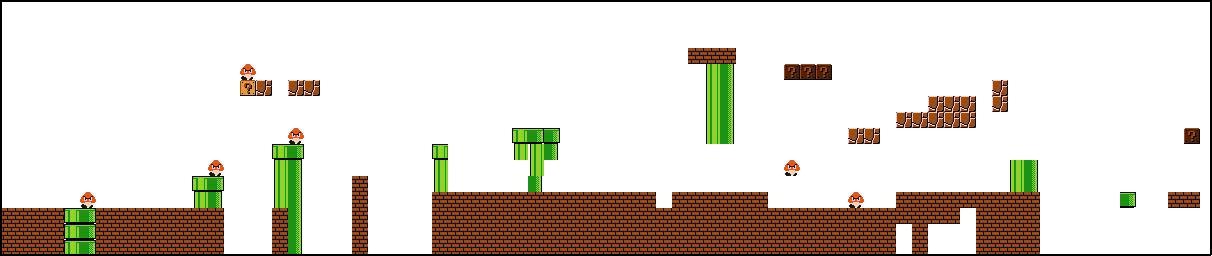}
            \caption{A defective level generated by GAN}
            \label{fig:GAN_start}
        \end{subfigure}
        \begin{subfigure}{.5\textwidth}
            \centering
            \includegraphics[width=1\textwidth]{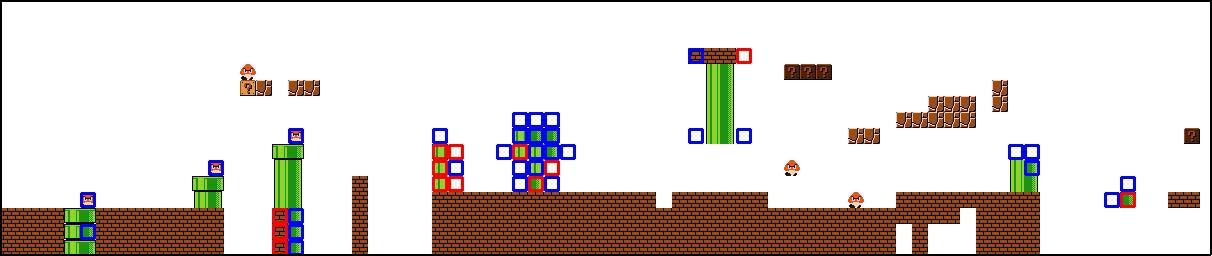}
            \caption{Wrong Tiles (Red) and Unstable Tiles (Blue) are marked by Cnet}
            \label{fig:GAN_start_mark}
        \end{subfigure}

        \begin{subfigure}{.5\textwidth}
            \centering
            \includegraphics[width=1\textwidth]{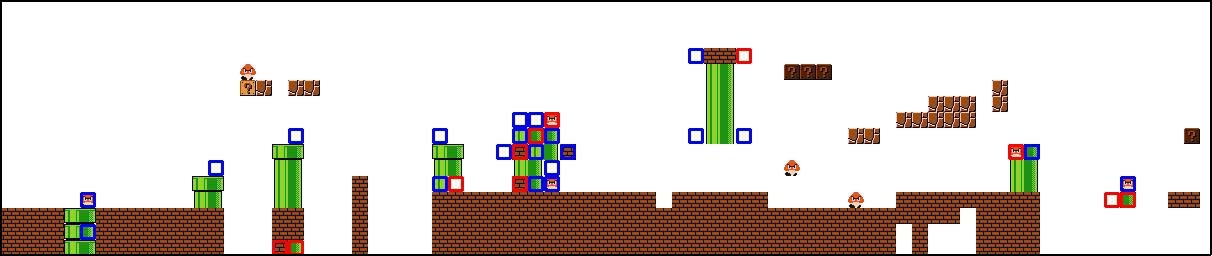}
            \caption{Best solution in the initialized population.}
            \label{fig:GAN_gen0}
        \end{subfigure}
        \begin{subfigure}{.5\textwidth}
            \centering
            \includegraphics[width=1\textwidth]{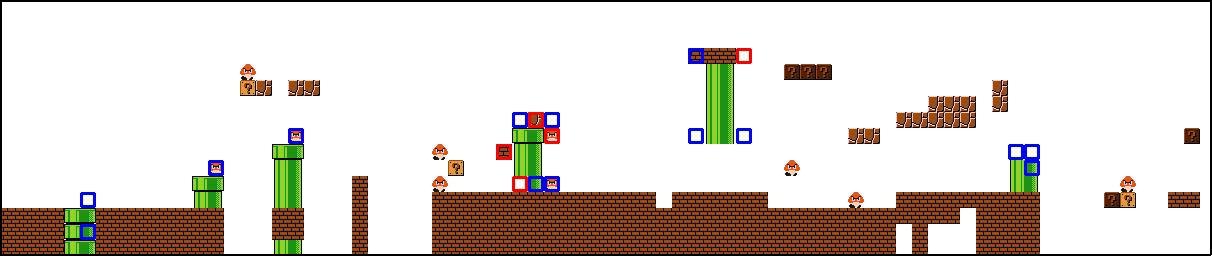}
            \caption{Best solution after 2 generations.}
            \label{fig:GAN_gen2}
        \end{subfigure}
        \begin{subfigure}{.5\textwidth}
             \centering
             \includegraphics[width=1\linewidth]{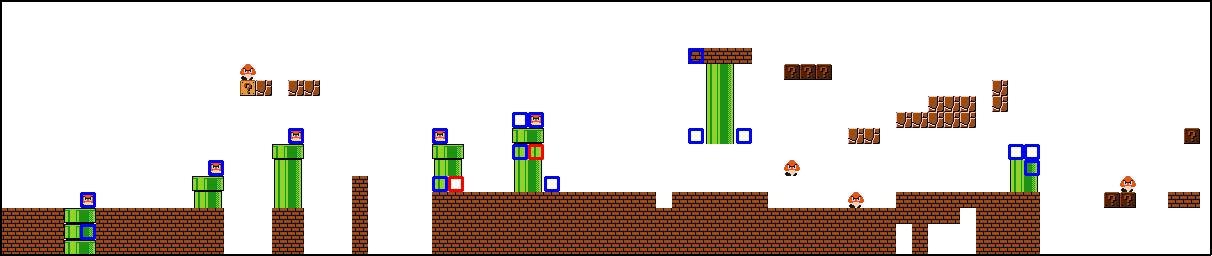}
             \caption{Best solution after 4 generations.}
             \label{fig:GAN_gen4}
         \end{subfigure}
                \begin{subfigure}{.5\textwidth}
            \centering
            \includegraphics[width=1\linewidth]{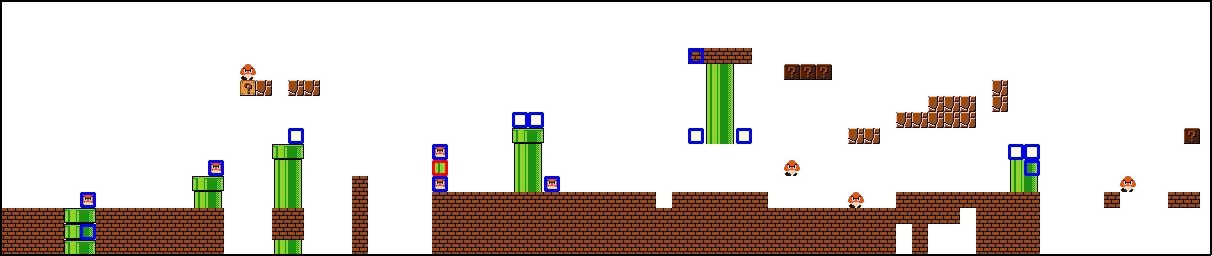}
            \caption{Best solution after 8 generations.}
            \label{fig:GAN_gen8}
        \end{subfigure}
        \begin{subfigure}{.5\textwidth}
            \centering
            \includegraphics[width=1\linewidth]{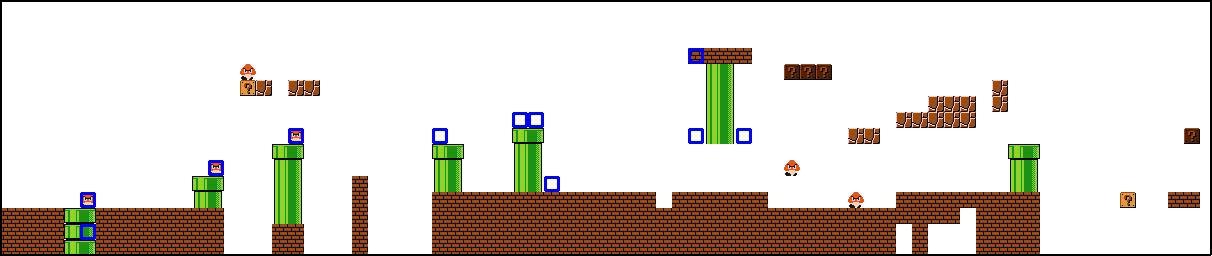}
            \caption{Best solution after 25 generations.}
            \label{fig:GAN_gen25}
        \end{subfigure}
    \caption{Repairing a defective level generated by GAN.}
    \label{fig:repair_gan}
    \end{figure}

\begin{figure*}[htbp]
        \centering
        \begin{subfigure}{1\textwidth}
            \centering
            \includegraphics[width=1\linewidth]{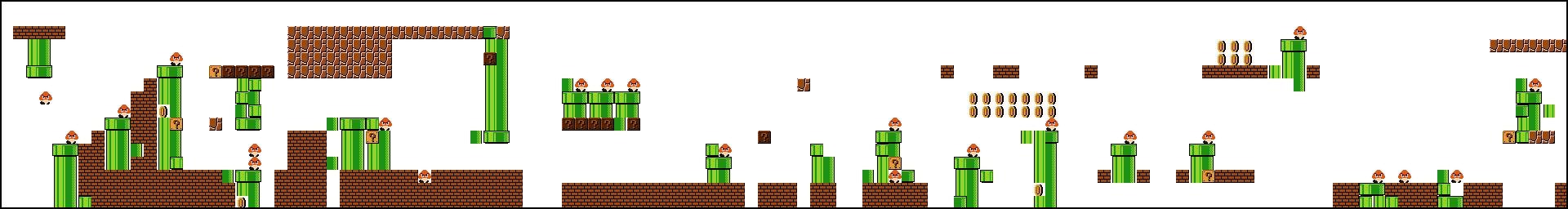}
            \caption{A randomly destroyed level}
            \label{fig:random_destroy_start}
        \end{subfigure}
        
        \begin{subfigure}{1\textwidth}
            \centering
            \includegraphics[width=1\linewidth]{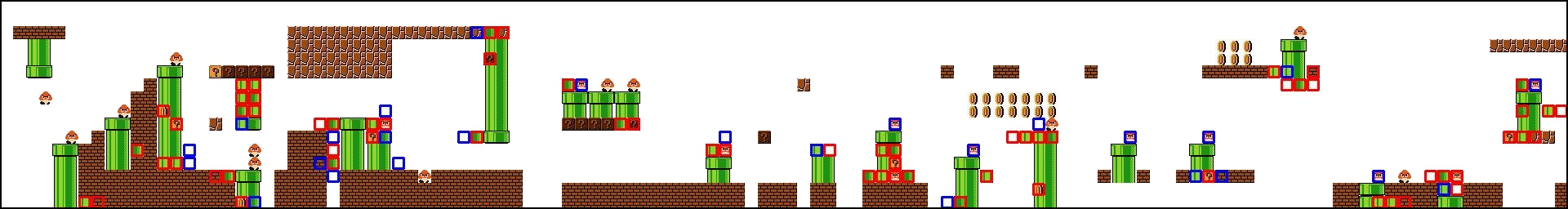}
            \caption{Wrong Tiles (Red) and Unstable Tiles (Blue) are marked by CNet}
            \label{fig:random_destroy_start_marked}
        \end{subfigure}
        
        \begin{subfigure}{1\textwidth}
            \centering
            \includegraphics[width=1\linewidth]{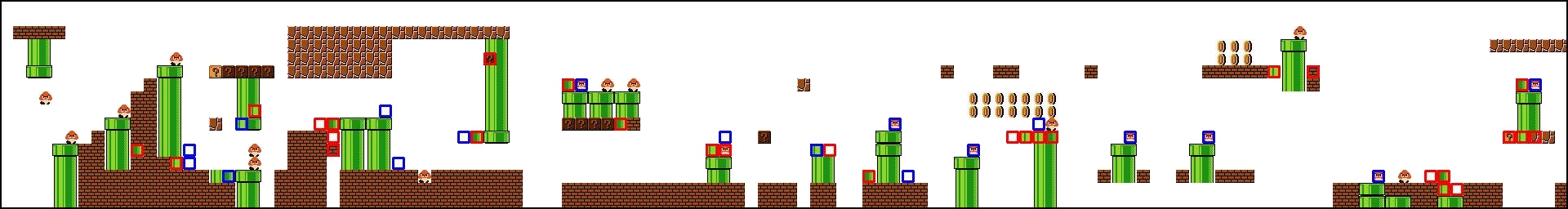}
            \caption{Best solution in the initialized population.}
            \label{fig:random_destroy_gen0}
        \end{subfigure}
        
        \begin{subfigure}{1\textwidth}
            \centering
            \includegraphics[width=1\linewidth]{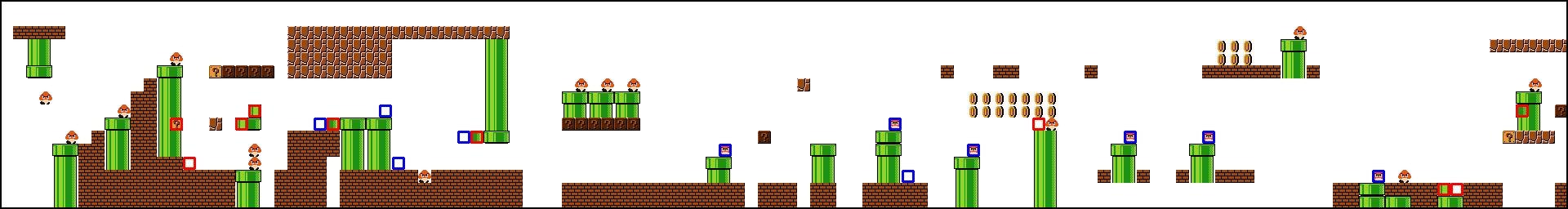}
            \caption{Best solution after 2 generations.}
            \label{fig:random_destroy_gen2}
        \end{subfigure}
        \begin{subfigure}{1\textwidth}
            \centering
            \includegraphics[width=1\linewidth]{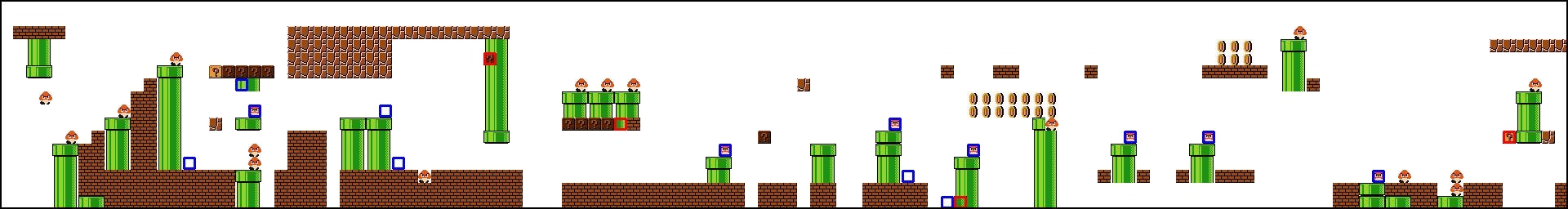}
            \caption{Best solution after 4 generations.}
            \label{fig:random_destroy_gen4}
        \end{subfigure}
        \begin{subfigure}{1\textwidth}
            \centering
            \includegraphics[width=1\linewidth]{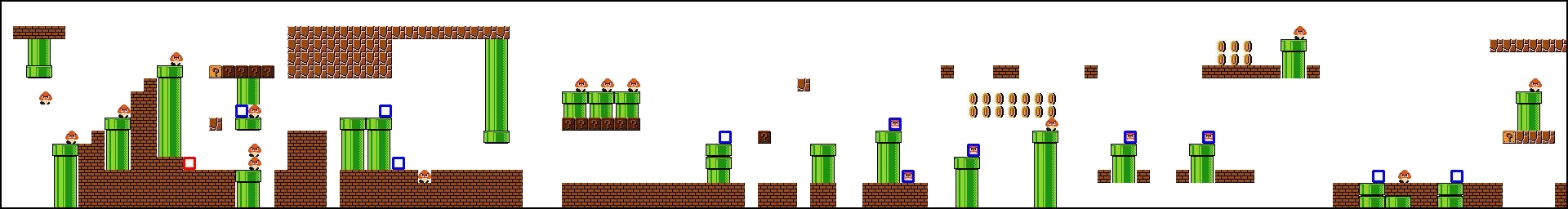}
            \caption{Best solution after 8 generations.}
            \label{fig:random_destroy_gen8}
        \end{subfigure}
        \begin{subfigure}{1\textwidth}
            \centering
            \includegraphics[width=1\linewidth]{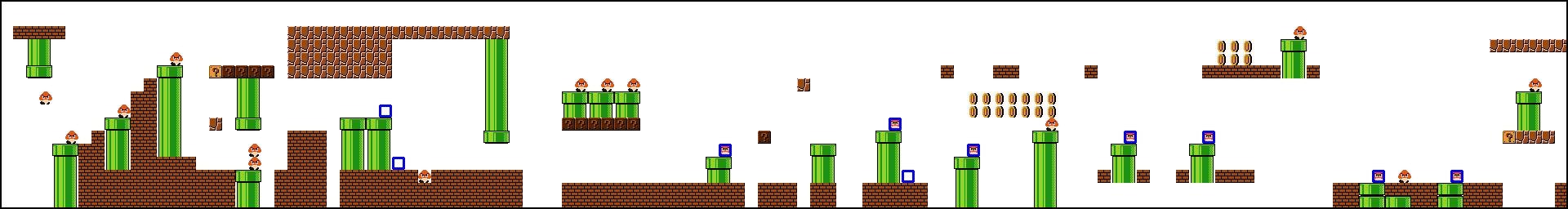}
            \caption{Best solution after 25 generations.}
            \label{fig:random_destroy_gen25}
        \end{subfigure}
        \caption{Repairing randomly destroyed levels. All kind of pipes appeared in the training data are collected in one figure.}
    \label{fig:random_destroy_repair}
    \end{figure*}

    We design two levels with flaw to test the performance of our evolutionary repairer. In the first test, we train a GAN with the same structure of the one described in \cite{volz2018evolving}, while the same training data as for training the CNet (detailed previously in Section \ref{sec:rulenet}). Then, some defective levels are collected from the levels generated by GAN (Fig. \ref{fig:GAN_start}). In the second test, different types of pipes from the training maps have been collected and put in one map. Then, tiles in this map are randomly selected to be destroyed. An example is shown in Fig. \ref{fig:random_destroy_start}. 
    
    In all the experiments, the weights in the fitness function (Eg. \ref{eq:fitness}) are set as $\omega_1 = 5 $, $\omega_2 = 3 $ and $\omega_3 = 1$ after fine tuning. All the individuals in the population are initialized by randomly setting a tile type for the unstable tiles in solution space (cf. Section \ref{sec:solutionspace}) using uniform distribution and applying repair once as described in Algorithm \ref{algo:operators}. The initial values of parameters are shown in the first line of Algorithm \ref{algo:evolution}. 
    
    When a tile is determined as a wrong tile, its alternative can be randomly selected from the Candidate Tiles following a designed probability distribution. 
    Three of the most common ways are using (i) a distribution with one element of 1 and others 0, thus winner-takes-all; (ii) a normalized distribution after truncating the probability output by the CNet or (iii) a uniform distribution. We have tested all of them in our experiments and only give the results of the third one in Figs. \ref{fig:GAN_start} and \ref{fig:random_destroy_start} due to its superior performance comparing to the other two and the length limit of the paper. 
    The failure of (i) has been explained previously in Section \ref{sec:recommender}.
    The failure of (ii) is probably due to the fact that empty tiles often occupy with the highest probability when surrounding info is fake.
    
    Fig. \ref{fig:repair_gan} illustrates an example of the progress of repairing a defective level generated by GAN. At the beginning (Fig. \ref{fig:GAN_start_mark}), there are some wrong tiles (highlighted by red boxes) and many unstable tiles (highlighted by blue boxes), due to the influence of fake surrounding info and the low truncation threshold $\theta$ (cf. Section \ref{sec:recommender}). When the optimization continues, the number of wrong tiles and unstable tiles is decreasing. After 4 generations only (Fig \ref{fig:GAN_gen4}), there are only 2 wrong tiles and a small amount of unstable tiles left. Then, after 8 generations (Fig \ref{fig:GAN_gen8}), only one wrong tile remains. After 25 generations, the level is almost perfectly repaired. Compared with the original level, the flaws are repaired perfectly and only a few tiles were changed. It is consistent with the design of the fitness function. Some broken pipes are erased, while some are repaired. This case study shows that genetic algorithm can get a variety of reasonable results.
    
    Fig. \ref{fig:random_destroy_repair} shows the results of repairing randomly destroyed levels. From Fig. \ref{fig:random_destroy_start_marked}, we can see that destroyed tiles are well recognized and marked by our CNet. Fig. \ref{fig:random_destroy_gen25} shows the best repaired level after 25 generations. We can see that all different broken pipes are well repaired. However, a few tiles around pipes are changed. There are two possible reasons: the evolutionary process is uncertain; the CNet has not been trained perfectly and may mark some legal tiles as wrong tiles.
    
\begin{figure}[htbp]
    \centering
    \includegraphics[width=0.8\linewidth]{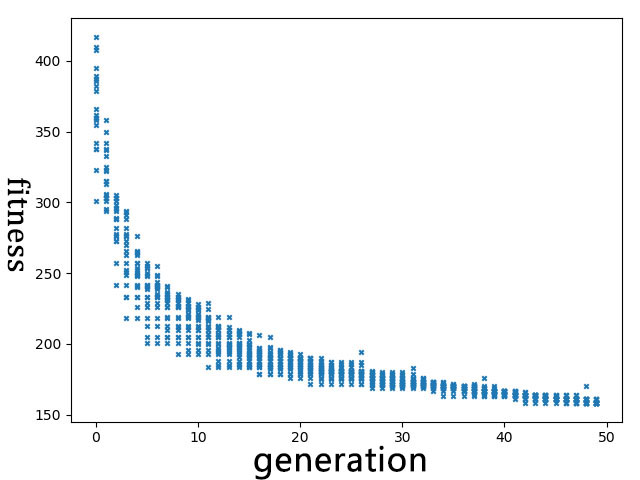}
        \caption{Scatter diagram of the fitness of all individuals in an arbitrary evolution process.}
    \label{fig:evolutionPoint}
\end{figure}

\begin{figure}[htbp]
    \centering
    \includegraphics[width=0.8\linewidth]{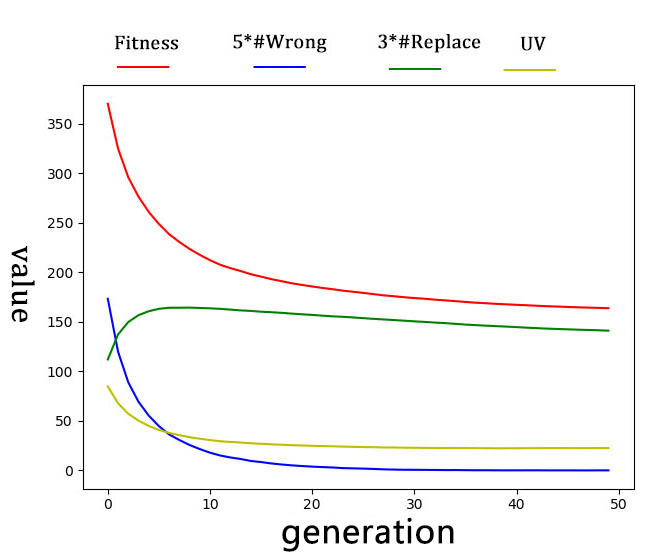}
        \caption{Evolutionary curves of the fitness and its items among the population, averaged over 30 trials, respected to generation.}
    \label{fig:evolutionLine}
\end{figure}

    To visualize the evolution process, we present the evolutionary curves of the values of fitness and the three items in the fitness function (Eq. \eqref{eq:fitness}) among the population averaged over 30 trials in Fig. \ref{fig:evolutionLine}. The Unstable Value (yellow curve) and the number of Wrong tiles (blue curve) decrease rapidly. This is because of the repair step described in Algorithm \ref{algo:operators}. The number of replaced tiles increases slowly. When it stops increasing around generation 5, the other values still decreases and the number of replaced tiles takes a main part of the fitness value, as what we expected, replacing fewer tiles while repairing the level better. Fig. \ref{fig:evolutionPoint} is the scatter diagram of all individuals when repairing a randomly destroyed level in an arbitrary optimization run. The fitness values of individuals converge gradually and the diversity among individuals' fitness is clearly shown.

    \subsection{Discussion}
    We design an evolutionary repairer to search for optimal replacement schemes for repairing defective levels. This evolutionary repairer is equipped with a novel search space being constructed with the help of CNet, and a novel fitness function considering the number of wrong tiles, the number of replaced tiles and unstable tiles in a level. The proposed approach is proved to be effective in our experiments of repairing GAN-generated Mario levels (Fig. \ref{fig:repair_gan}) and randomly destroyed levels (Fig. \ref{fig:random_destroy_start}).
    \section{Statistical Significance and Analysis of Repaired Levels}\label{sec:statistical}
    To examine the statistical significance of our CNet and evolutionary repairer, we trained 10 CNet models using distinct random seeds which were further used to repair 10 randomly destroyed levels generated from the level with all types of pipes designed in Section \ref{sec:gaxps}.
    Each model has been trained for 4,000 epochs and uses more strict criteria to evaluate repair results. Tiles whose surrounding info appear (height is not included) in training data are labelled as right, otherwise, the tiles are wrong. Batch size is set as 1. We compare the tiles in the level segments before and after repairing, and analyse the number of tiles that have been edited and their status in Table \ref{table_10net}. 
    
\begin{figure}
    \centering
    \includegraphics[width=0.3\textwidth]{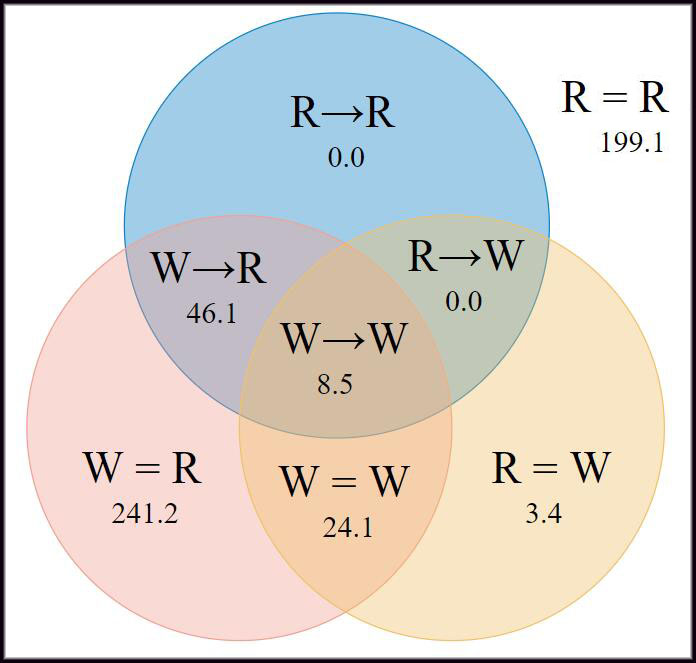}
    \caption{The venn diagram of the last column in Table \ref{table_10net}. The universal set represents all tiles which contains pipes in surrounding info. The red set contains tiles which are wrong before repair. The yellow set contains tiles which are wrong after repair. The blue set contains tiles which are replaced during repair. $R=R$ represents the complementary set of the union of the blue, red and yellow sets.}
    \label{fig:venn}
\end{figure}

    The notations used in Table \ref{table_10net} are defined as follows. The symbols ``$\rightarrow$'' and ``$=$'' indicate if a tile is replaced or not after repair. ``W'' and ``R'' indicate if a tile is labelled as wrong or right. Therefore, $W\rightarrow W$ refers to the situation that a wrong tile is replaced but remains being labelled as wrong; $W=R$ refers to the situation that a tile is not replaced, but its label is changed from wrong to right after being repaired due to the changes of it's surrounding tiles; $W=W$ means that a wrong tile is not replaced at all. $R\rightarrow R$, $R=R$, $R\rightarrow W$, $W\rightarrow R$ are defined similarly. The bottom row illustrates the ratio between the amount of wrong tiles after and before repair, thus the smaller the better. A venn diagram (Fig. \ref{fig:venn}) is also included to assist the understanding of relations between these changes of tile status. 
    
    We can see that the amount of R$\rightarrow$R and R$\rightarrow$W is zero which means our method will not replace the right tiles. The yellow set is very small compared with the red set. It means our method can fix most of wrong tiles. W=W takes a big part of the yellow set. The reason is that our CNet is not trained perfectly. Our CNet will regard some wrong tiles as right tiles. So these tiles are not replaced during evolution process and remain wrong. So the performance of our method is based on the quality of CNet a lot. Avoiding overfitting and training a better CNet will decrease the wrong tiles after repair.

\begin{table}[htbp]
    \setlength{\tabcolsep}{1pt}
    \centering
        \caption{\label{table_10net}Repair results of 10 CNet models trained with distinct random seeds. The original level segment with all pipe types has been randomly destroyed to generate 10 broken level segments, which have been repaired individually using the evolutionary repairer described in Section \ref{sec:enhancedga}. Each cell shows the result averaged over the 10 levels fixed by each CNet model. The notations are explained in Section \ref{sec:statistical}.}
    \begin{tabular}{c|rrrrrrrrrr|r}
    \toprule
    Before  & \multicolumn{10}{c|}{Index of Net} &  \multirow{2}{*}{Avg.}\\
      /After   & 1 & 2 & 3 & 4 & 5 & 6 & 7 & 8 & 9 & 10 &\\
    \midrule
W$\rightarrow$W & 8.8 & 8.9 & 8.3 & 7.7 & 7.1 & 9.0 & 8.5 & 10.6 & 9.4 & 7.0 & 8.5 \\
W$\rightarrow$R & 46.7 & 45.9 & 46.4 & 46.4 & 47.1 & 45.6 & 46.2 & 44.3 & 45.2 & 46.9 & 46.1 \\
R$\rightarrow$W & 0.0 & 0.0 & 0.0 & 0.0 & 0.0 & 0.0 & 0.0 & 0.0 & 0.0 & 0.0 & 0.0 \\
R$\rightarrow$R & 0.1 & 0.0 & 0.0 & 0.1 & 0.0 & 0.0 & 0.1 & 0.0 & 0.1 & 0.0 & 0.0 \\
W=W & 23.1 & 26.6 & 23.7 & 22.0 & 18.8 & 22.8 & 26.1 & 31.1 & 27.0 & 20.2 & 24.1 \\
W=R & 241.3 & 238.5 & 241.5 & 243.8 & 246.9 & 242.5 & 239.1 & 233.9 & 238.3 & 245.8 & 241.2 \\
R=W & 3.3 & 4.5 & 4.1 & 3.2 & 2.2 & 2.2 & 4.1 & 4.5 & 3.6 & 2.2 & 3.4 \\
R=R & 199.2 & 198.0 & 198.4 & 199.3 & 200.3 & 200.3 & 198.4 & 198.0 & 198.9 & 200.3 & 199.1 \\
\hline
Ratio ($\%$) & 6.5 & 7.5 & 6.7 & 6.0 & 5.1 & 6.3 & 7.3 & 8.8 & 7.5 & 5.4 & 6.7 \\
    \bottomrule
    \end{tabular}
\end{table}
\section{Conclusion and future work}\label{sec:conclusion}

    In the last few years, more and more PCGML methods have been applied to level generation. However, logic errors may occurs in the generated levels. Therefore, we focus on video game levels that can be represented by 2D images and design a CNet and an evolutionary repairer for automatically detecting errors in levels generated by PCGML and repairing the defective levels. 
    
    In this work, we consider a specific case: broken pipes in levels of \smb. 
    We design a CNet to detect and repair error tiles in Mario levels, and an evolutionary repairer with a special search space and novel heuristic to search for replacement schemes used for repairing the defective levels. Our CNet-assisted evolutionary repairer by combining these two approaches successfully identified the error tiles in levels generated by MarioGan~\cite{volz2018evolving} and repaired nearly without specific human knowledge for this game in our experiments. This CNet-assisted evolutionary repairer can be used together with the MarioGan~\cite{volz2018evolving} for designing and repairing levels for other games which can be encoded in the similar way.
    

      
    Our approaches can be further improved in different ways. The performance of our CNet-assisted evolutionary repairer highly relies on the quality of the trained CNet model.
    Improving the structure or the training process of CNet to make it more reliable is worth investigating. What's more, our CNet was tested on learning the constraints of pipes in \smb only. Training CNets for all the constraints with the whole set of 35 tile types in \smb is challenging but worthy of study in the future.
    Another future work is applying our approaches to other games of which the levels can be represented as images, such as The Legend of Zeld and Angry Birds, and to generating landscapes for 3D games.

 \section*{Acknowledgment}
The authors would like to thank the authors of \cite{volz2018evolving} for making their MarioGAN framework open-source. The authors would like to thank the Dagstuhl Seminar 19511.

\bibliographystyle{IEEEtran}
\balance
\bibliography{main.bib}

\def\toolong{
\appendix

\section{Appendix}
Figs. \ref{fig:example1} and \ref{fig:example2} illustrate 10 repaired Mario level segments with all types of pipes by a CNet model. As shown in this 10 examples, most of the broken pipes can be fixed by our CNet.

\begin{figure*}[htbp]
    \begin{subfigure}{0.8\textwidth}
        \centering
        \includegraphics[width=1\linewidth]{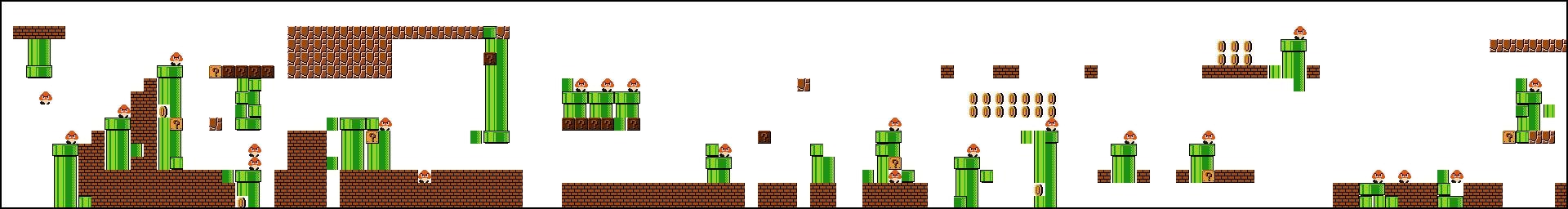}
        \caption{Damaged segment 1.}
        \label{fig:dagamed_lvl1}
    \end{subfigure}
    \centering
    \begin{subfigure}{0.8\textwidth}
        \centering
        \includegraphics[width=1\linewidth]{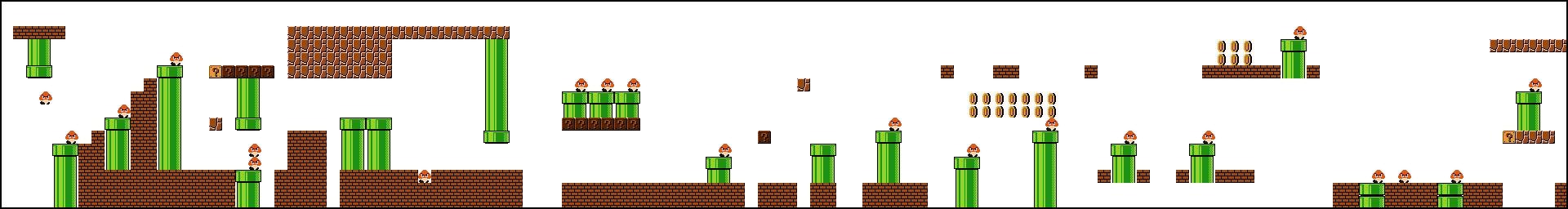}
        \caption{Repaired segment 1.}
        \label{fig:repaired_lvl1}
    \end{subfigure}
    \centering
    \begin{subfigure}{0.8\textwidth}
        \centering
        \includegraphics[width=1\linewidth]{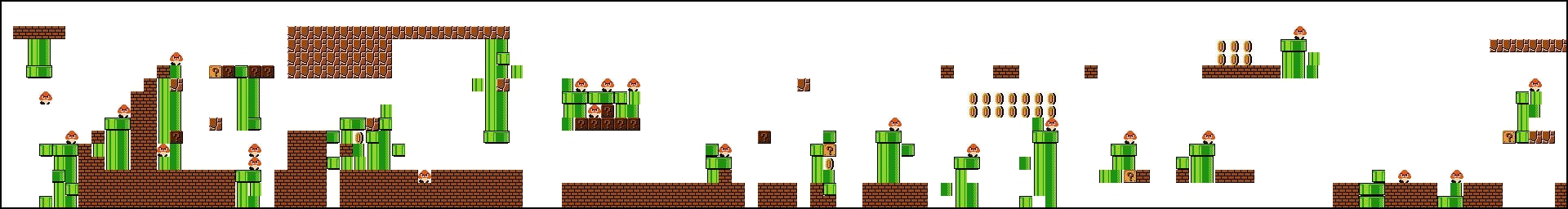}
        \caption{Damaged segment 2.}
        \label{fig:dagamed_lvl2}
    \end{subfigure}
    \centering
    \begin{subfigure}{0.8\textwidth}
        \centering
        \includegraphics[width=1\linewidth]{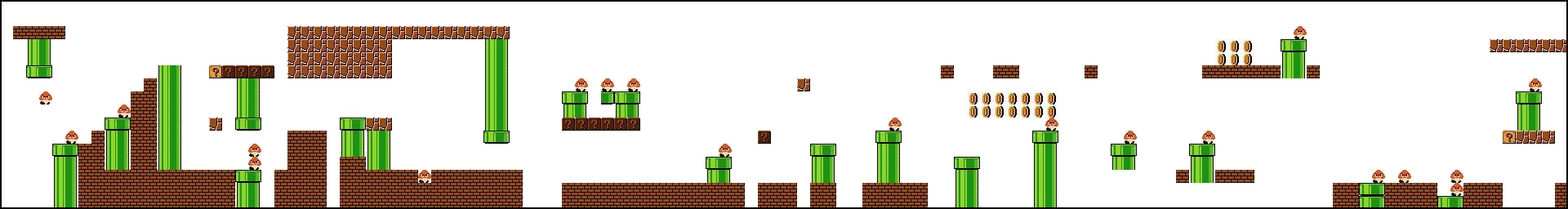}
        \caption{Repaired segment 2.}
        \label{fig:repaired_lvl2}
    \end{subfigure}
    \centering
    \begin{subfigure}{0.8\textwidth}
        \centering
        \includegraphics[width=1\linewidth]{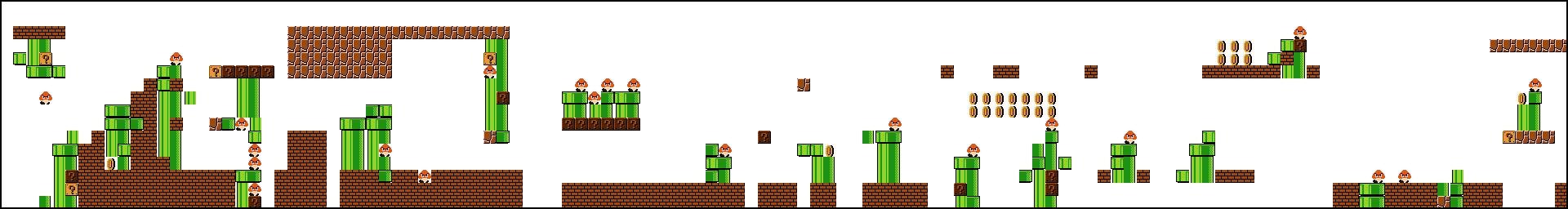}
        \caption{Damaged segment 3.}
        \label{fig:dagamed_lvl3}
    \end{subfigure}
    \centering
    \begin{subfigure}{0.8\textwidth}
        \centering
        \includegraphics[width=1\linewidth]{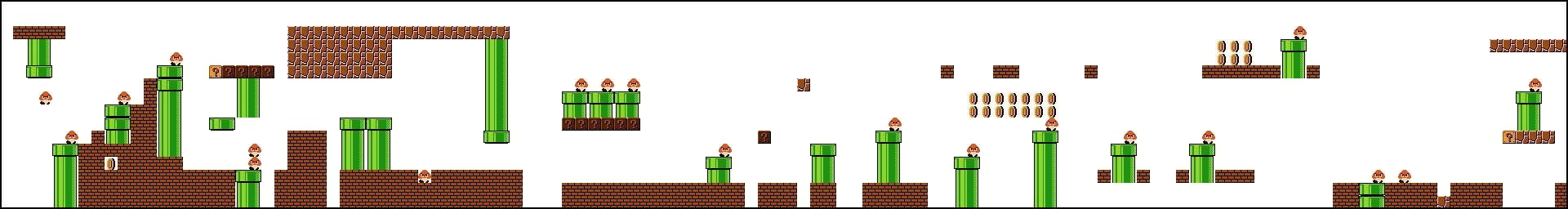}
        \caption{Repaired segment 3.}
        \label{fig:repaired_lvl3}
    \end{subfigure}
    \centering
        \begin{subfigure}{0.8\textwidth}
        \centering
        \includegraphics[width=1\linewidth]{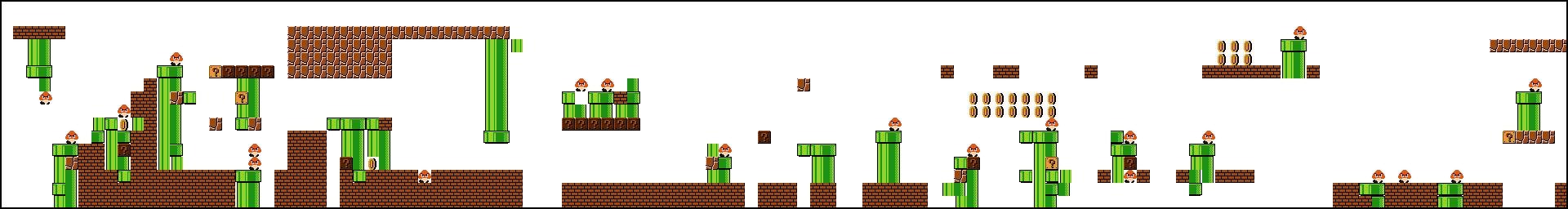}
        \caption{Damaged segment 4.}
        \label{fig:dagamed_lvl4}
    \end{subfigure}
    \centering
    \begin{subfigure}{0.8\textwidth}
        \centering
        \includegraphics[width=1\linewidth]{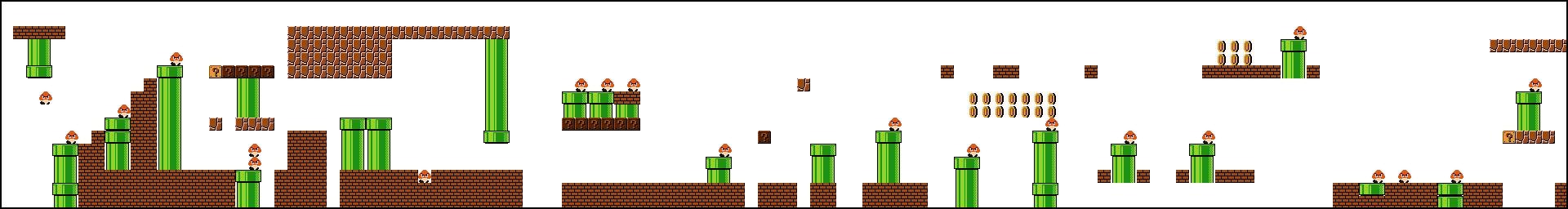}
        \caption{Repaired segment 4.}
        \label{fig:repaired_lvl4}
    \end{subfigure}
    \centering
        \begin{subfigure}{0.8\textwidth}
        \centering
        \includegraphics[width=1\linewidth]{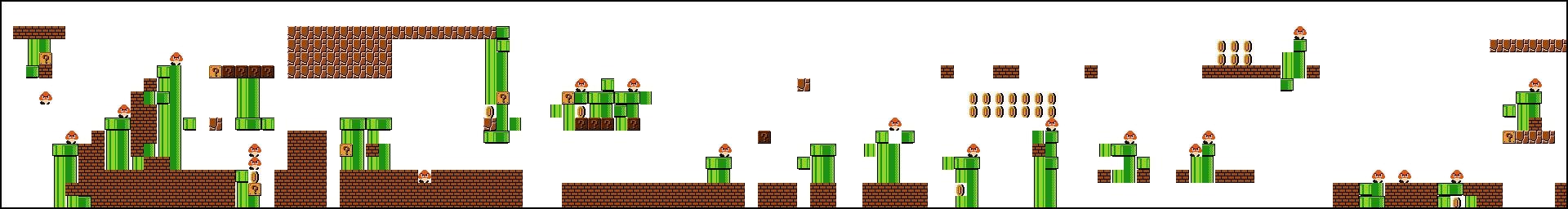}
        \caption{Damaged segment 5.}
        \label{fig:dagamed_lvl5}
    \end{subfigure}
    \centering
    \begin{subfigure}{0.8\textwidth}
        \centering
        \includegraphics[width=1\linewidth]{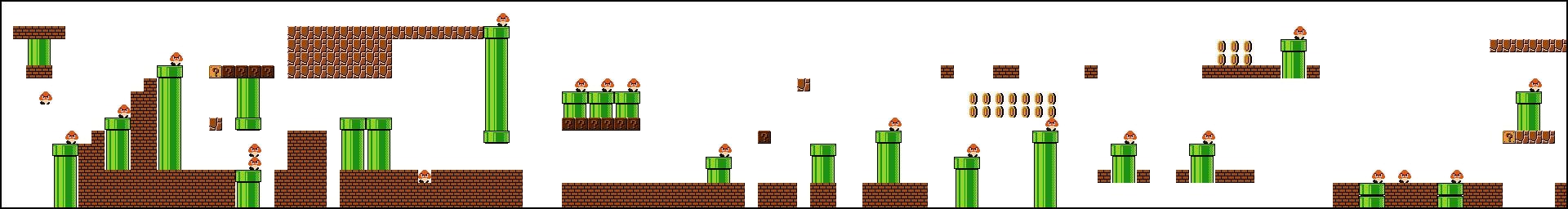}
        \caption{Repaired segment 5.}
        \label{fig:repaired_lvl5}
    \end{subfigure}
  \caption{\label{fig:example1}Illustration of repairing 10 damaged level segments with all types of pipes by a CNet (to be continued in Fig. \ref{fig:example2}).}
\end{figure*}

\begin{figure*}[htbp]
    \centering
    \begin{subfigure}{0.8\textwidth}
        \centering
        \includegraphics[width=1\linewidth]{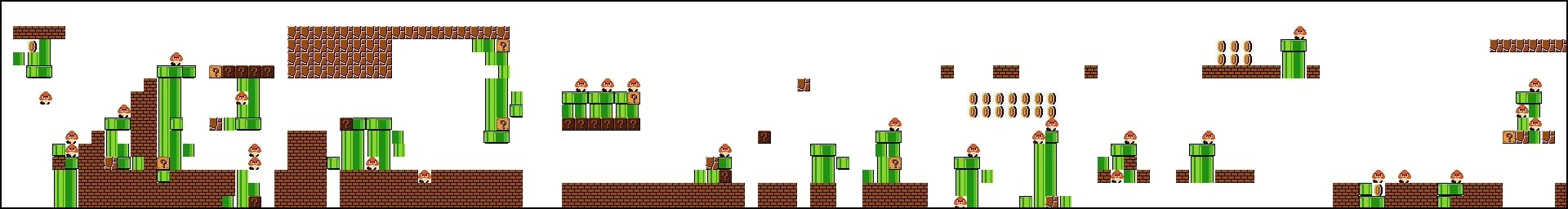}
        \caption{Damaged segment 6.}
        \label{fig:dagamed_lvl6}
    \end{subfigure}
    \centering
    \begin{subfigure}{0.8\textwidth}
        \centering
        \includegraphics[width=1\linewidth]{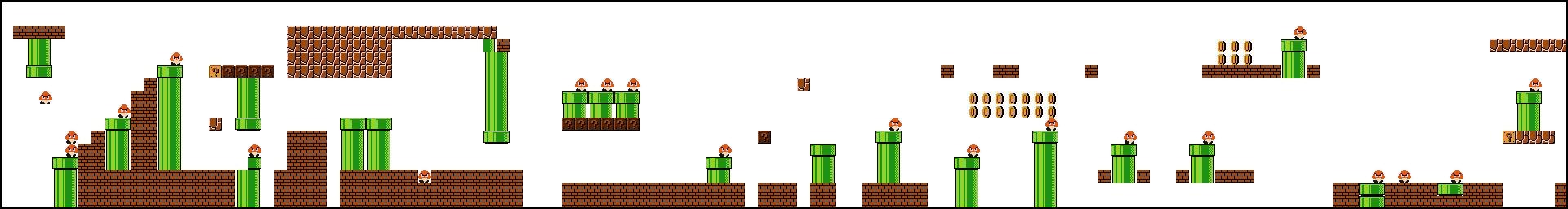}
        \caption{Repaired segment 6.}
        \label{fig:repaired_lvl6}
    \end{subfigure}
    \centering
        \begin{subfigure}{0.8\textwidth}
        \centering
        \includegraphics[width=1\linewidth]{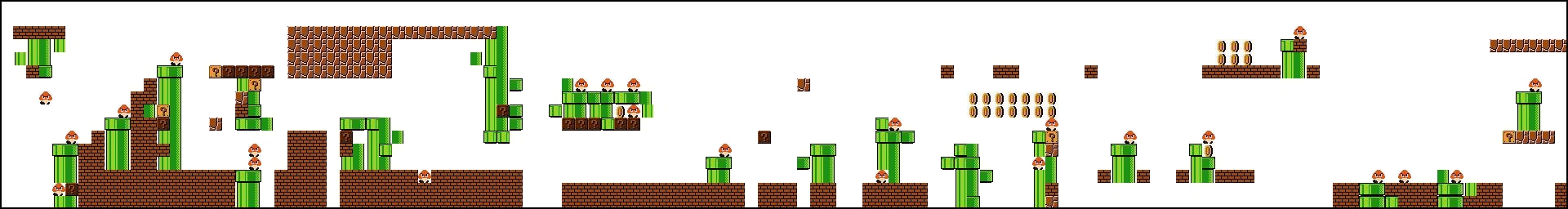}
        \caption{Damaged segment 7.}
        \label{fig:dagamed_lvl7}
    \end{subfigure}
    \centering
    \begin{subfigure}{0.8\textwidth}
        \centering
        \includegraphics[width=1\linewidth]{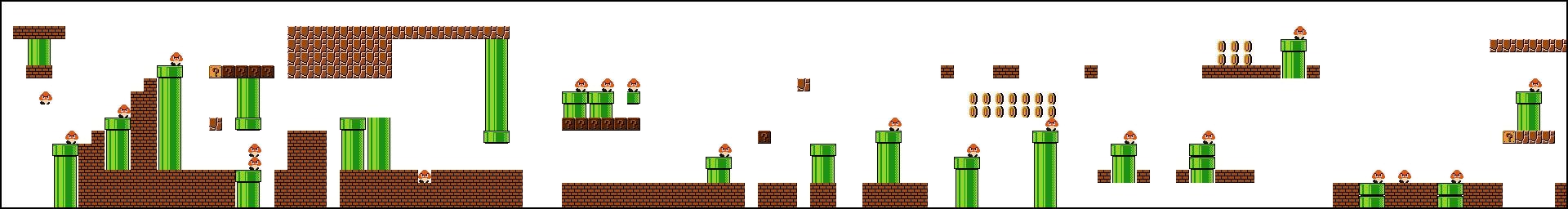}
        \caption{Repaired segment 7.}
        \label{fig:repaired_lvl7}
    \end{subfigure}
    \centering
        \begin{subfigure}{0.8\textwidth}
        \centering
        \includegraphics[width=1\linewidth]{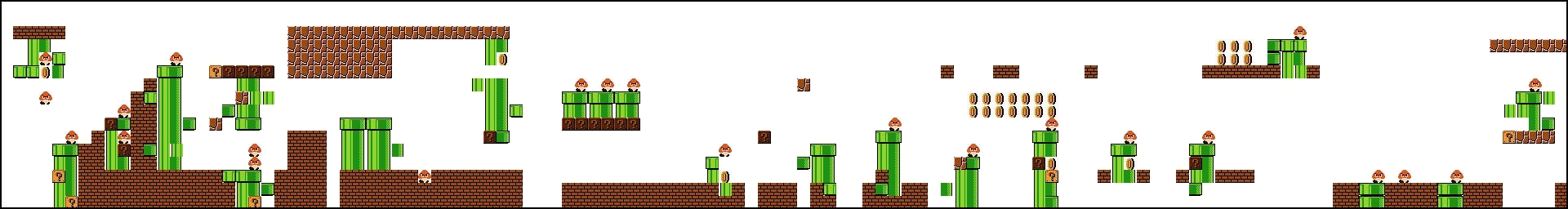}
        \caption{Damaged segment 8.}
        \label{fig:dagamed_lvl8}
    \end{subfigure}
    \centering
    \begin{subfigure}{0.8\textwidth}
        \centering
        \includegraphics[width=1\linewidth]{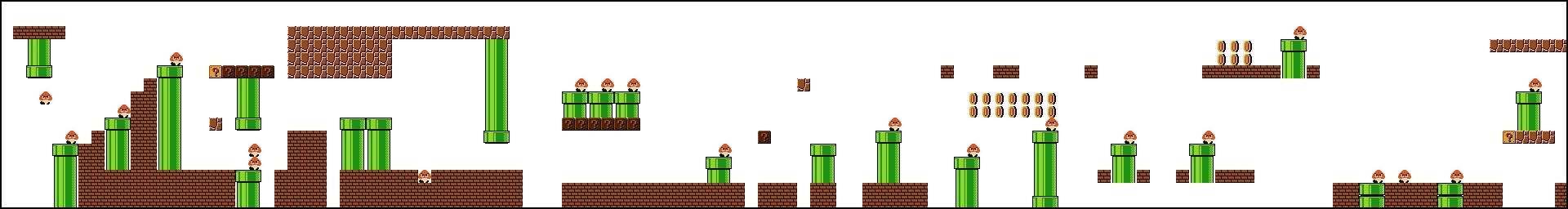}
        \caption{Repaired segment 8.}
        \label{fig:repaired_lvl8}
    \end{subfigure}
    \centering
        \begin{subfigure}{0.8\textwidth}
        \centering
        \includegraphics[width=1\linewidth]{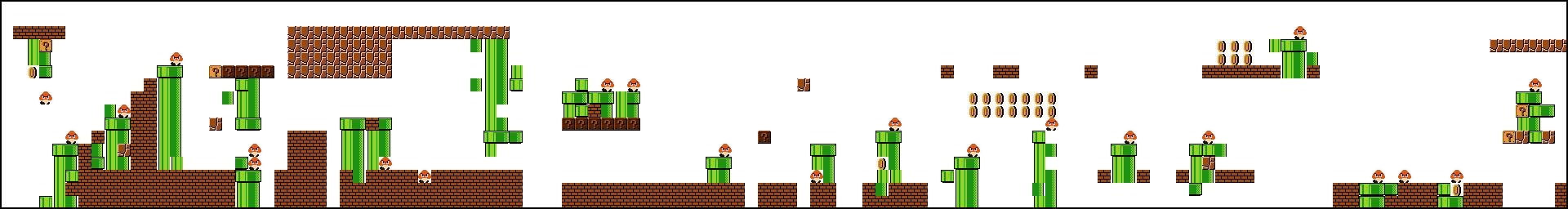}
        \caption{Damaged segment 9.}
        \label{fig:dagamed_lvl9}
    \end{subfigure}
    \centering
    \begin{subfigure}{0.8\textwidth}
        \centering
        \includegraphics[width=1\linewidth]{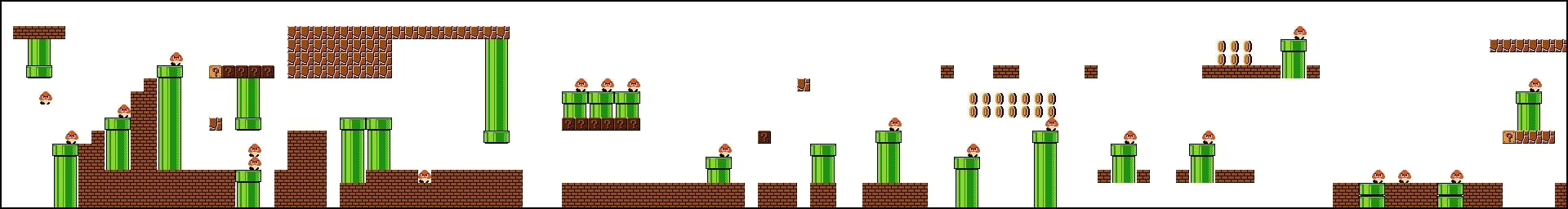}
        \caption{Repaired segment 9.}
        \label{fig:repaired_lvl9}
    \end{subfigure}
    \centering
        \begin{subfigure}{0.8\textwidth}
        \centering
        \includegraphics[width=1\linewidth]{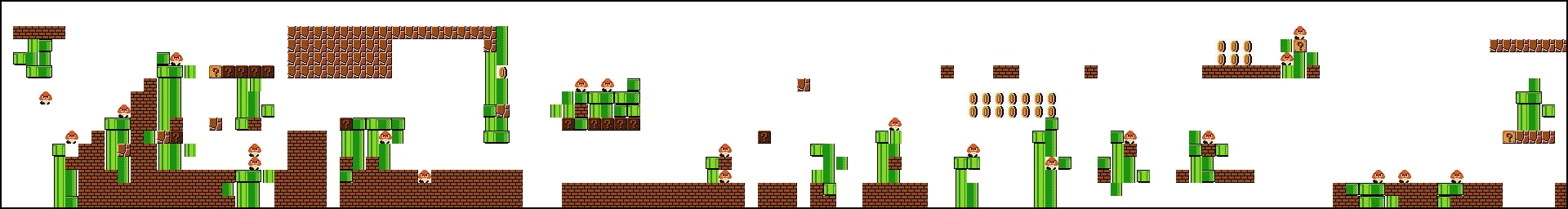}
        \caption{Damaged segment 10.}
        \label{fig:dagamed_lvl10}
    \end{subfigure}
    \centering
    \begin{subfigure}{0.8\textwidth}
        \centering
        \includegraphics[width=1\linewidth]{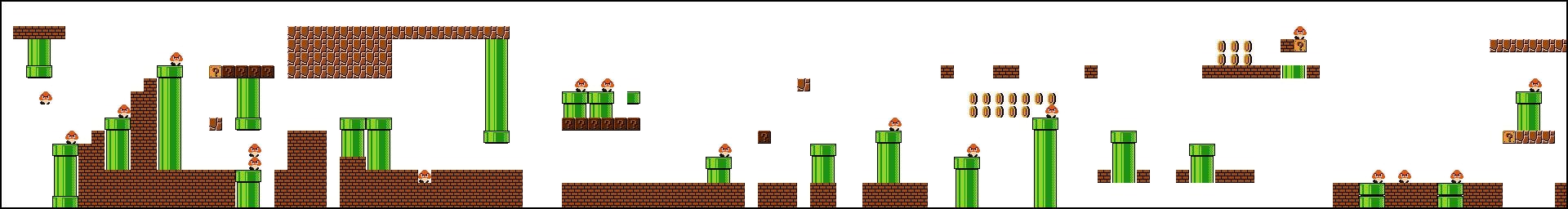}
        \caption{Repaired segment 10.}
        \label{fig:repaired_lvl10}
    \end{subfigure}
    \centering
    \caption{\label{fig:example2}(Continuation of Fig. \ref{fig:example1}) Illustration of repairing 10 damaged level segments with all types of pipes by a CNet.}
\end{figure*}
}
\end{document}